\pdfoutput=1
\documentclass[letterpaper, 10 pt, conference]{ieeeconf}

\IEEEoverridecommandlockouts
\usepackage{cite}
\usepackage{amsmath,amssymb,amsfonts}
\usepackage{graphicx}
\usepackage{textcomp}
\usepackage{xcolor}
\usepackage{amsthm}
\usepackage{caption}
\usepackage{subcaption}
\usepackage{hyperref}
\usepackage{todonotes} % 用来添加TODO
\usepackage{stfloats} % 双栏图片位置
\usepackage{algorithmic} % 伪代码
\usepackage[ruled]{algorithm2e} 
\usepackage{multirow} % 表格合并单元格
\usepackage{booktabs} % *号
\usepackage{threeparttable} % 带注释的表格
\hypersetup{ % 设置链接样式
    colorlinks = true,
    linkcolor=blue,
    filecolor=blue,
    urlcolor=blue,
    citecolor=cyan,
}
\usepackage{balance} % 最后一页文献对齐
\def\BibTeX{{\rm B\kern-.05em{\sc i\kern-.025em b}\kern-.08em
    T\kern-.1667em\lower.7ex\hbox{E}\kern-.125emX}}

\title{\LARGE \bf
Toward Global Sensing Quality Maximization: 

A Configuration Optimization Scheme for Camera Networks
}

\author{
Xuechao Zhang$^*$, Xuda Ding$^*$, Yi Ren$^\dagger$, Yu Zheng$^\dagger$, Chongrong Fang$^*$ and Jianping He$^*$
\thanks{
This work is supported by the CIE-Tencent Robotics X Rhino-Bird Focused Research Program. The authors would like to thank Dr. Xiaoming Duan, Han Wang and Hao Jiang for their helpful advice on technical issues.}
\thanks{$^*$: the Dept. of Automation, Shanghai Jiao Tong University, Shanghai, China. email: {\tt\small\{zhang.xc, dingxuda, crfang, jphe\}@sjtu.edu.cn}.
}
\thanks{$^\dagger$: the Tencent Robotics X Lab, Shenzhen, China. email: {\tt\small \{evanyren, petezheng\}@tencent.com}}
}

\begin{document}

\maketitle
\thispagestyle{empty}
\pagestyle{empty}

%%%%%%%%%%%%%%%%%%%%%%%%%%%%%%%%%%%%%%%%%%%%%%%%%%%%%%%%%%%%%%%%%%%%%%%%%%
\begin{abstract}
The performance of a camera network monitoring a set of targets depends crucially on the configuration of the cameras. 
In this paper, we investigate the reconfiguration strategy for the parameterized camera network model, with which the sensing qualities of the multiple targets can be optimized globally and simultaneously.
We first propose to use the number of pixels occupied by a unit-length object in image as a metric of the sensing quality of the object, which is determined by the parameters of the camera, such as intrinsic, extrinsic, and distortional coefficients. Then, we form a single quantity that measures the sensing quality of the targets by the camera network. This quantity further serves as the objective function of our optimization problem to obtain the optimal camera configuration. We verify the effectiveness of our approach through extensive simulations and experiments, and the results reveal its improved performance on the AprilTag detection tasks. Codes and related utilities for this work are open-sourced and available at https://github.com/sszxc/MultiCam-Simulation.

\end{abstract}
% \begin{IEEEkeywords}
% Camera Network System, Optimization, Multi-camera Sensing Fusion.
% \end{IEEEkeywords}
%%%%%%%%%%%%%%%%%%%%%%%%%%%%%%%%%%%%%%%%%%%%%%%%%%%%%%%%%%%%%%%%%%%%%%%%%%

% \listoftodos % 所有TODO

\section{Introduction}
% 现在相机很多-多相机网络的优势 但是需要配置
\subsection{Background and motivation}
% 各种算法对相机的感知质量有要求，覆盖率和质量有权衡，资源浪费，需要算法来配置（尤其是当目标变多的时候）
% 和覆盖不同，相机的感知质量主要受到多方面影响，衡量起来很复杂

\begin{figure}[t]
    \centering
    \includegraphics[width=0.7\linewidth]{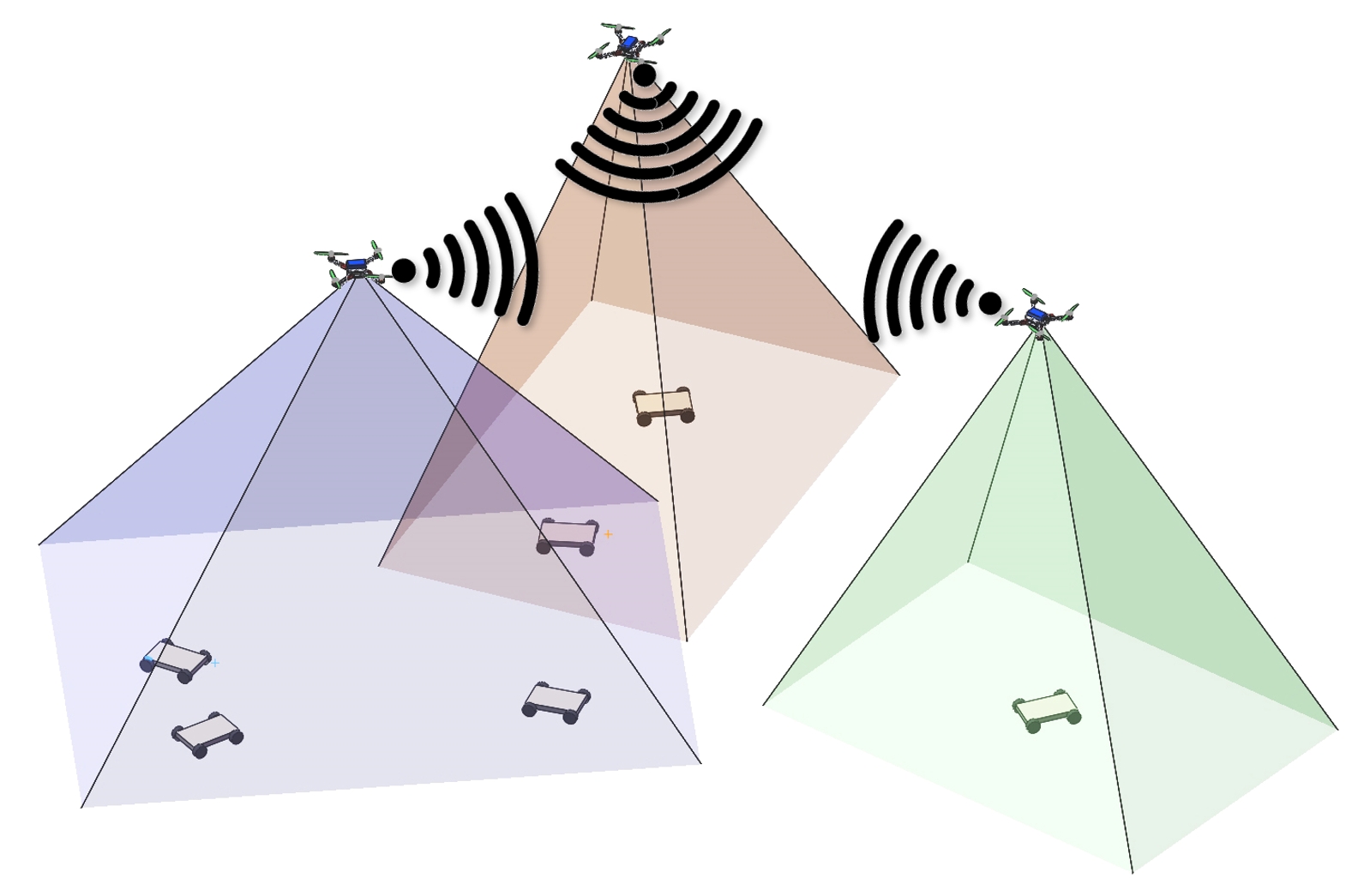}
    \caption{A network of three cameras mounted on the drones for tracking autonomous ground vehicles.}
    \label{fig:application scenario}
    % \vspace{-10pt}
\end{figure}

Cameras play important roles in society surveillance \cite{chen2021automated}, environmental monitoring \cite{bernhard2021case}, and scene perception, localization and mapping in robotics \cite{wu2018image}, etc. Compared to single-camera systems, camera networks tend to have a larger field of view and more viewing angles for each object, leading to stronger anti-interference ability and better detection performance \cite{yang2020multi,yang2017multi}.
Fig. \ref{fig:application scenario} shows a typical application for camera networks as drones tracking ground vehicles with cameras.
A typical task of camera networks is to provide adequate coverage for the targets\cite{zhang2018visual, suresh2020maximizing}. However, meeting the qualitative coverage requirement does not necessarily ensure the desirable sensing quality of the targets in the scene. The distortion inevitably introduced in the imaging process of cameras depends on the lens's physical properties, the camera's parameters, viewing angle and the objects' relative position. Different configurations of the camera network may result in different sensing quality of the targets.

Configuring a camera network properly to cover multiple targets while ensuring desirable sensing quality is challenging due to the vast amount of camera parameters involved and a lack of widely accepted sensing performance metric \cite{horster2006optimal}. In this paper, we propose a straightforward metric that captures the coverage and sensing quality of objects for a camera network, based on which we automate the search for the configuration that optimizes the proposed metric.

\subsection{Related Work}
% 过去的工作包括
% 1. 只考虑覆盖的一般 NP-hard，复杂情况只能启发式算法求解
% 2. 主观评价质量的，使用 voronoi
% 3. 需要提前建模、完整的仿真的一篇

Recent surveys \cite{mavrinac_modeling_2013}\cite{7095560} summarize the historical developments in the area of camera network coverage. Globally optimal configuration for such systems, similar to the \emph{Art Gallery Problem} \cite{o1987art} and \emph{Watchmen Tour Problem} \cite{efrat2000sweeping}, has been proved to be NP-hard, and using general optimization methods could lead to sub-optimal solutions. Recent literature reports several approaches that have been proposed to solve this problem. 

In \cite{konda_optimal_2013}\cite{konda_global_2016}, the authors focus on complex indoor environments and present a method for pan-tilt-zoom (PTZ) camera reconfiguration. The proposed method considers the targets, camera distortion, and environment illumination, and a particle swarm optimizer gives the solution. However, the performance of the method relies on complete modeling of the environment, which is generally very difficult to obtain.
The authors in \cite{aghajanzadeh_camera_2020} take into consideration the realistic constraints of computer vision and aim to achieve a balance between coverage and resolution for multiple cameras. They split the 2D plane into polygons with equal areas and use a greedy algorithm to segment the polygons for a simplified camera model and determine the cameras' configuration.

Recently, a class of gradient methods based on Voronoi diagrams has been widely used to solve the problem of sensor allocation. In \cite{arslan_voronoi-based_2018}, the authors focus on a 2D convex environment and propose a sensing quality metric using prior knowledge of the imaging process. They successfully use a provably correct greedy algorithm to configure cameras for given event distribution.

% 少一个总结

\subsection{Contributions and Organization}
% 我们的工作贡献：新问题、新模型、实验
% 文章规划
In this paper, we build an automatic reconfiguration system for camera networks that significantly improves collective sensing performance. The main contributions are as follows.
\begin{itemize}
\item We propose a new metric that measures the coverage and sensing quality of the targets for a camera. This metric is easy to compute and is shown to be effective in simulation and experiments.
\item We propose a novel model to quantitatively describe the effect of the camera distortion on sensing quality, which is suitable for any pre-calibrated cameras and targets in 3D space.
\item We verify the reliability and usability of the strategy through sufficient experiments both in virtual and physical environments.
\end{itemize}

The rest of the paper is organized as follows. Section \ref{sec:Formulation} explains the relationship between sensing quality and the number of pixels occupied by the target, and presents the proposed sensing quality model for a multi-camera system. In Section \ref{sec:Experimental Validation}, we illustrate the effectiveness of the proposed reconfiguration model using simulations and experimental results, respectively. Section \ref{sec:Conclusion} concludes with a summary of our contributions and a brief discussion of future directions.

\section{Problem Statement and Formulation}\label{sec:Formulation}
% 说明感知质量和占据像素数量之间的关系
% 除了分辨率、焦距等固有属性 影响因素只有透视和畸变
% 说明透视和畸变都可以用这种方法描述
% 所以这是一个好模型

% Calibration model from https://docs.opencv.org/3.0-beta/modules/calib3d/doc/camera_calibration_and_3d_reconstruction.html#
% some symbol definition refers to \url{https://zhuanlan.zhihu.com/p/94244568}

\begin{figure}[t]
	\centering
	\includegraphics[width=0.6\linewidth]{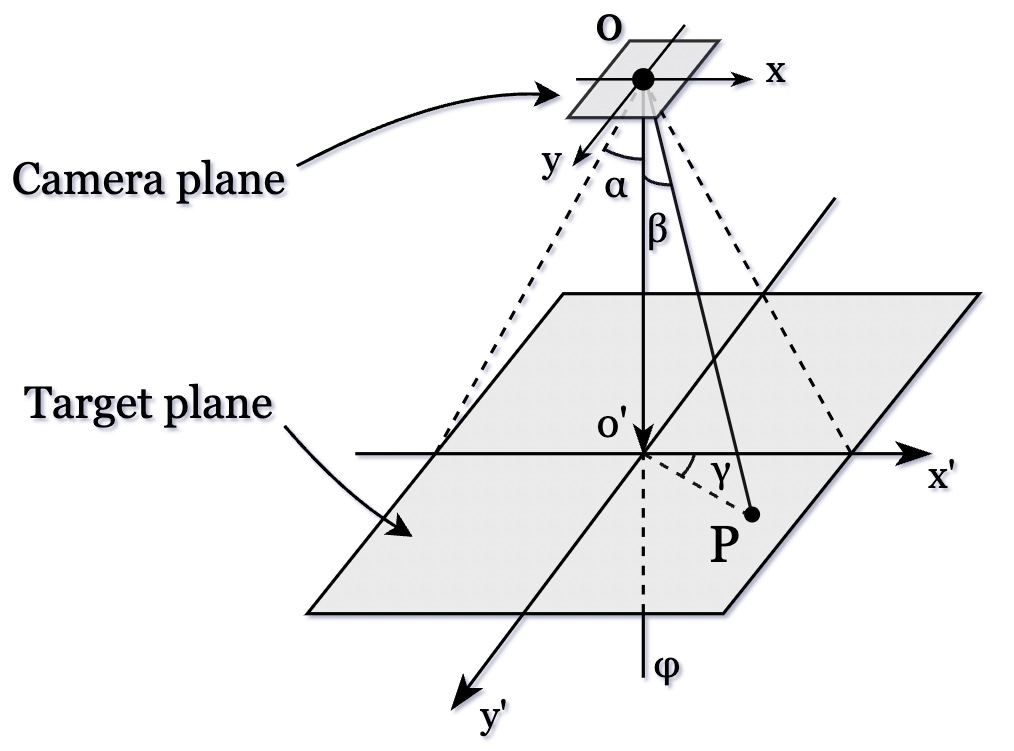}
	\caption{Camera symbol definition.}
    \label{fig:camera_symbol}
    \vspace{-1pt}
\end{figure}

A general camera network consists of several heterogeneous cameras that may have distinct properties. Dynamic configuration adjustment of the camera network must be performed collectively to ensure that the entire network of cameras is globally optimized concerning the specified task or performance requirements. The sensing process is camera-specific and mutually exclusive, but when the cameras' fields of view overlap the fusion results should be calculated jointly.

As shown in Fig. \ref{fig:camera_symbol}, we consider a model for a camera $C_i\in \mathbb{C}$ for $i\in\{1,2,\ldots,N_c\}$ with a conic field of view, whose configuration is specified by its fixed position $ o_i \in \mathbb{R}^3 $, variable optical-axis direction $\phi_i \in \mathbb{R}^3$, angle of view $ 2\alpha_i \in (0,\pi) $ and maximum radial resolution of $w_i$ pixels. The position of the target $T_j \in \mathbb{T}$ for $ j\in\{1,2,\ldots,N_t\}$ in the sensing task is denoted by $p_j \in \mathbb{R}^3$. 

Let $o'_i$ be the point such that $\overrightarrow{o'_ip_j} \perp \phi_i$. The plane that is parallel to the camera plane and contains $o'_i$ is the target plane. Let $\beta^i_j$ be the angle between $\overrightarrow{o_ip_j}$ and $\phi_i$, and $\gamma^i_j$ be the angle between $\overrightarrow{o'_ip_j}$ and $\overrightarrow{o'_ix'}$. If $\hat{p_j}$ is the estimated position for $T_j$, then the sensing error is defined by
\begin{equation}
\epsilon_j = \left \| p_j-\hat{p}_j \right \| _2.
\end{equation}

Counting the number of pixels a target occupies in the image before correcting distortion is an intuitive description of the sensing quality. To illustrate the idea, suppose the target with the length of $l$ occupies $n$ pixels in the image.
% suppose the length of a target is $l$, and it occupies $n$ pixels in the image. 
Moreover, let the actual length represented by each pixel on the target plane be $k$. Then the measurement $l'$ has an error $|l-l'| < 2k$. The reason is that each pixel on complementary metal–oxide–semiconductor (CMOS) measures the average brightness information through integration\cite{kuroda2017essential}, and the error caused by sampling will make both ends of the target captured as an uncertain pixel. It shows that when the same target occupies more pixels, the sensor has richer information to describe the target, making the downstream object detection more accurate. In other words, a shorter actual length represented by each pixel results in a smaller relative error. The error of positioning $\epsilon_j$
is bounded by
\begin{equation}
\epsilon_j < k.
\end{equation}

% \begin{figure}[ht]
% 	\centering
% 	\includegraphics[width=0.8\linewidth]{pic/model_pixel_error.jpg}
% 	\caption{pixel error}
% 	\label{fig:pixel_error}
% \end{figure}

% 分别介绍单相机的透视、畸变，多相机的融合，整个问题的约束
\subsection{Single-Camera Model}
This section discusses how the number of pixels occupied by the target as a metric of sensing quality is related to two critical parameters, i.e., the camera's perspective factor and the distortion factor.

\subsubsection{Perspective}
\begin{figure}[t]
	\centering
	\includegraphics[width=0.722\linewidth]{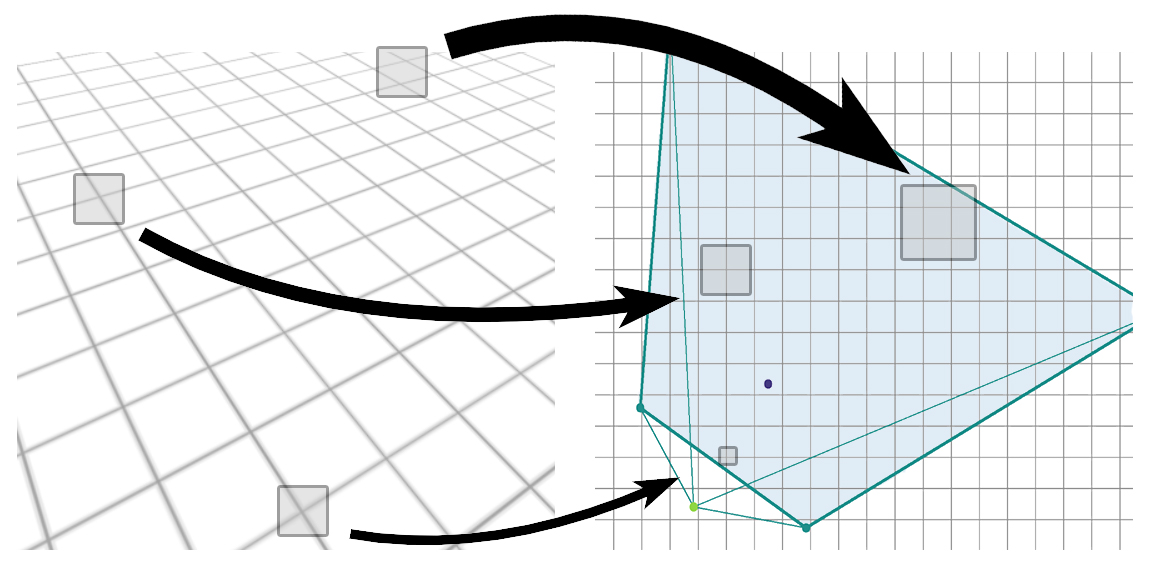}
	\caption{Perspective comparison: (left) image space and (right) world space. The farther away from the lens, the larger the area described by a single pixel, and vice versa.}
	\label{fig:perspective_principle}
	% \vspace{-9pt}
\end{figure}

Perspective primarily depends on the angle of view and focal length which considerably deteriorates the sensing quality. Objects appear smaller as their distance from the observer increases. In imaging process, the actual distance between the camera and the object has a great influence on the number of pixels that the object occupies, illustrated in Fig. \ref{fig:perspective_principle}. It is shown that the number of pixels occupied by an object with unit length is inversely proportional to the distance between the camera and the object. To improve sensing quality, the camera should be placed closer to the target.

Suppose that camera $ C_i $ performs the task of tracking target $ T_j $ and assume only perspective factor is considered. According to \cite{robinson1988optical}, the sensing quality here can be denote as
% \todo{weird?add more}
\begin{equation}
q_{p}\triangleq\frac{2|\overrightarrow{o_ip_j}|\cos\beta^i_j\tan\alpha_i}{w_i}.
\label{eq:kp}
\end{equation}

\subsubsection{Distortion}

Distortions can be introduced at the image acquisition and processing stages. With a specific lens and perspective, distortions are inherent in the captured videos which can seriously affect the achievable results \cite{bisagno_virtual_2018}. In the pinhole camera model system, a widely-used mathematical formulation to correct camera distortions is the Brown-Conrady model \cite{fryer1986lens} which includes radial and slight tangential distortion.
\begin{equation}
\begin{split}
& x'= xK_r + 2s_1xy+s_2(r^2+2x^2), \\
& y'= yK_r + s_1(r^2+2y^2)+2s_2xy,
\end{split}
\label{eq:distortion}
\end{equation}
where $r^2=x^2+y^2$ and
\begin{equation}
    K_r = \frac{1+k_1r^2+k_2r^4+k_3r^6}{1+k_4r^2+k_5r^4+k_6r^6}.
\end{equation}

In \eqref{eq:distortion}, $ (x,y) $ and $ (x',y') $ stands for the points before and after the distortion, respectively. $k_1, k_2, k_3, k_4, k_5,$ and $k_6$ are radial distortion coefficients. $s_1$ and $s_2$ are tangential distortion coefficients. These coefficients can be easily calculated by using classical calibration algorithms\cite{zhang1999flexible}.

\begin{figure}[t]
    \centering
    \includegraphics[width=0.4\textwidth]{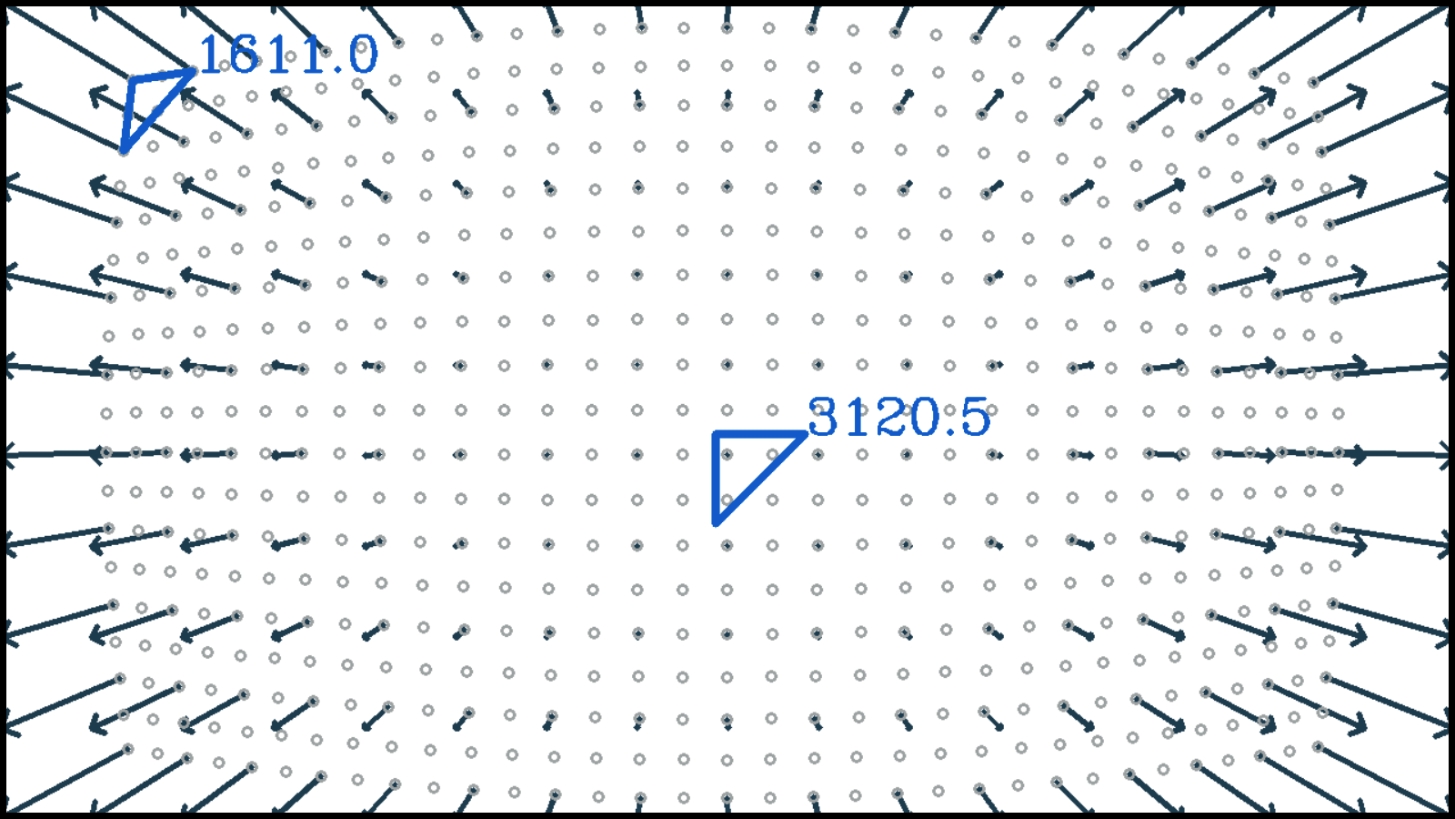}
    \caption{The arrows indicate the mapping relationship of the pixels during the distortion correction, and the blue triangles represent the origin areas on the sensor occupied by two triangles of same size after correction. In this example, there is 48.4\% less information at the edge of the image than in the center.}
    \label{fig:distort}
    \vspace{-5pt}
\end{figure}
\begin{figure*}[b]
% \vspace{-10pt}
	\centering
	\begin{subfigure}{0.25\linewidth}
		\centering
		\includegraphics[width=0.9\linewidth]{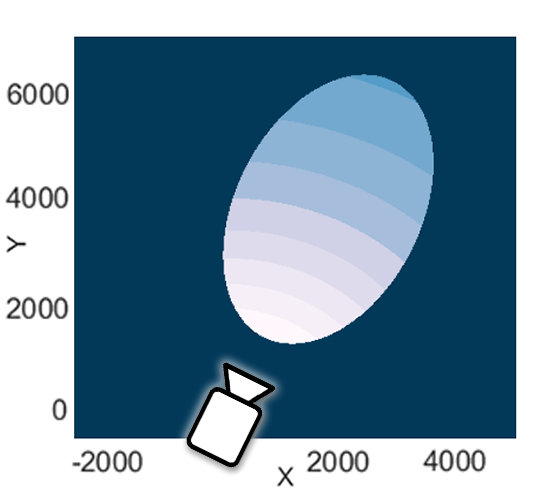}
		\caption{Perspective factor}
	\end{subfigure}
	\centering
	\begin{subfigure}{0.25\linewidth}
		\centering
		\includegraphics[width=0.9\linewidth]{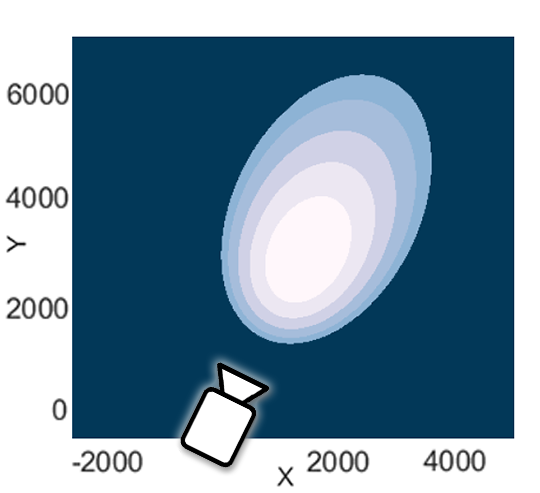}
		\caption{Distortion factor}
	\end{subfigure}
	\centering
	\begin{subfigure}{0.25\linewidth}
		\centering
		\includegraphics[width=0.9\linewidth]{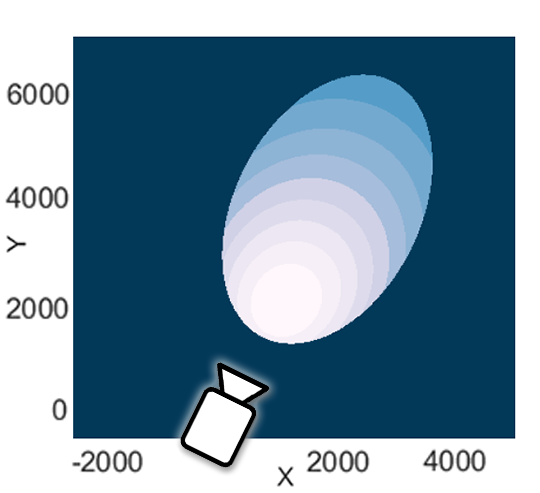}
		\caption{Complete model}
	\end{subfigure}
	\centering
	\begin{subfigure}{0.04\linewidth}
		\centering
		\includegraphics[width=0.9\linewidth]{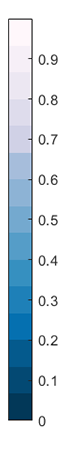}
	\end{subfigure}
	\caption{The sensing quality on the ground plane by a certain camera mounted at (0, 0, 3000)mm. The XY axis represents the position, and the lighter the color, the higher the quality. 
% 	(a) Perspective factor. (b) Distortion factor. (c) Complete model.
    }
	\label{fig:camera_model}
\end{figure*}
% To perform tasks such as positioning, distortion correction is necessary. 
Distortion correction is necessary for performing detection, positioning, and recognition tasks.
The distortion correction process of an image is a non-linear mapping, which does not bring any new information about the targets as the differences in pixel density are compensated by interpolation algorithms. After the stretching, the effective information at the edge and the center of the corrected image has a huge gap, as shown in Fig. \ref{fig:distort}.

Furthermore, the model coefficients for distortion correction can also be used to evaluate the effective pixel density obtained by the sensor during the imaging process. The pixels are stretched in different proportions in the directions of x and y. Denote $x_t', x_{t+1}'$ as two adjacent pixels after distortion correction, then $|x_t - x_{t+1}|$ represents the original information provided by the sensor on x direction. According to \eqref{eq:distortion}, the influence of distortion factor can be written as the derivative of $x', y'$. The pixel density is the product of the distortion factor in the directions of x and y. Therefore, for camera $C_i$ and target $T_j$, we use $q_{d}$ to denote the influence factor of distortion on pixel density.
\begin{equation}
\begin{split}
    &x'=\frac{\tan\beta^i_j\cos\gamma^i_j}{\tan\alpha_i}, \\
    &y'=\frac{\tan\beta^i_j\sin\gamma^i_j}{\tan\alpha_i}, \\
    &q_{d}\triangleq\frac{\partial x'}{\partial x}\frac{\partial y'}{\partial y}.
\end{split}
\label{eq:qd}
\end{equation}

Thus, the influence of different types of lenses can be described uniformly, no matter how they distort lights. This constraint indicates that the camera should be configured to align its optical axis with the object.

Finally, combining \eqref{eq:kp} and \eqref{eq:qd} yields a comprehensive description of the sensing quality for a single camera as
\begin{equation}
Q_j=q_{p}q_{d} > \epsilon_j.
\label{eq:complete_model}
\end{equation}
An illustration of this model is shown in Fig. \ref{fig:camera_model}.

% \begin{remark}
% It is worth mentioning that $k_{p}$ and $q_{d}$ both describe the actual length represented by a unit pixel and the multiplication in \eqref{eq:complete_model} retains their ... meaning perfectly(?) when taking both factors into account. Since the perspective and distortion factor exist in all types of lenses, this metric can be a universe model with strong applicability ...
% \end{remark}

\begin{figure}[t]
	\centering
	\includegraphics[width=0.75\linewidth]{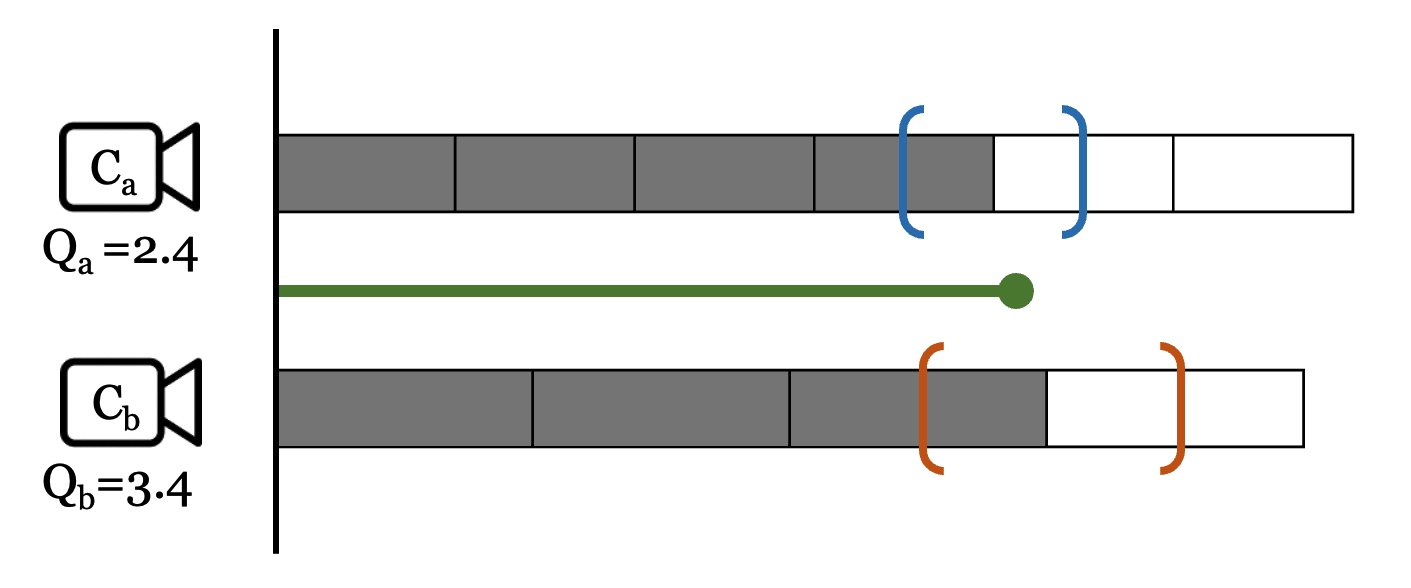}
	\caption{A one-dimensional example for fusion: The task is to estimate the length of the green line, and the fused result is the intersection of two cameras with different $Q_j$.}
    \label{fig:fusion}
    \vspace{-10pt}
\end{figure}

\subsection{Multi-Camera Model}

To extend the single-camera model to multiple cameras, some modifications need to be made to take into account the information fusion. Now consider two cameras $C_a,C_b$ observe the same target, for example, and $Q_a,Q_b$ (let $Q_a<Q_b$) denote the actual length represented by the unit pixel at the target location of the two cameras. Fusing data from the two cameras allows for more precise measurements than sensing with a single camera.
More specifically, the fused result is the intersection of the origin range, that is $\epsilon_j \in [0,Q_a]$ considering all possibilities of each camera's relative position and focal length. % Assume that the coefficients of the camera and the position of the object appear randomly (uniformly distributed?), 
A one-dimensional example for intuitive demonstration is given in Fig. \ref{fig:fusion}. Suppose that the overlap between cameras occurs randomly. The expectation of the error bound can be written as

\begin{equation}
\overline{Q}_j = \frac{1}{1/Q_a+1/Q_b}.
\end{equation}
% = \frac{2Q^a*Q^a/2+(Q^b-Q^a)*Q^a}{Q^a+ Q^b}

Similarly, it can be extended to the camera network of arbitrary scale. For a camera $C_i$ and a target $j$, use $V_{ij}$ to denote the visibility.
\begin{equation}
V_j^i = \left\{
\begin{aligned}
1, \beta^i_j \leq \alpha_i \\
0, \beta^i_j > \alpha_i.
\end{aligned}
\right.
\end{equation}
Then, the fusion of every cameras' estimation is written as

\begin{equation}
\overline{Q}_j = \frac{1}{ \sum_{C_i \in \mathbb{C}} V_j^i/Q_j^i} > \epsilon_j.
\label{eq:fusion result}
\end{equation}
\begin{figure} [b]
\vspace{-15pt}
    \centering
    \includegraphics[width=1\linewidth]{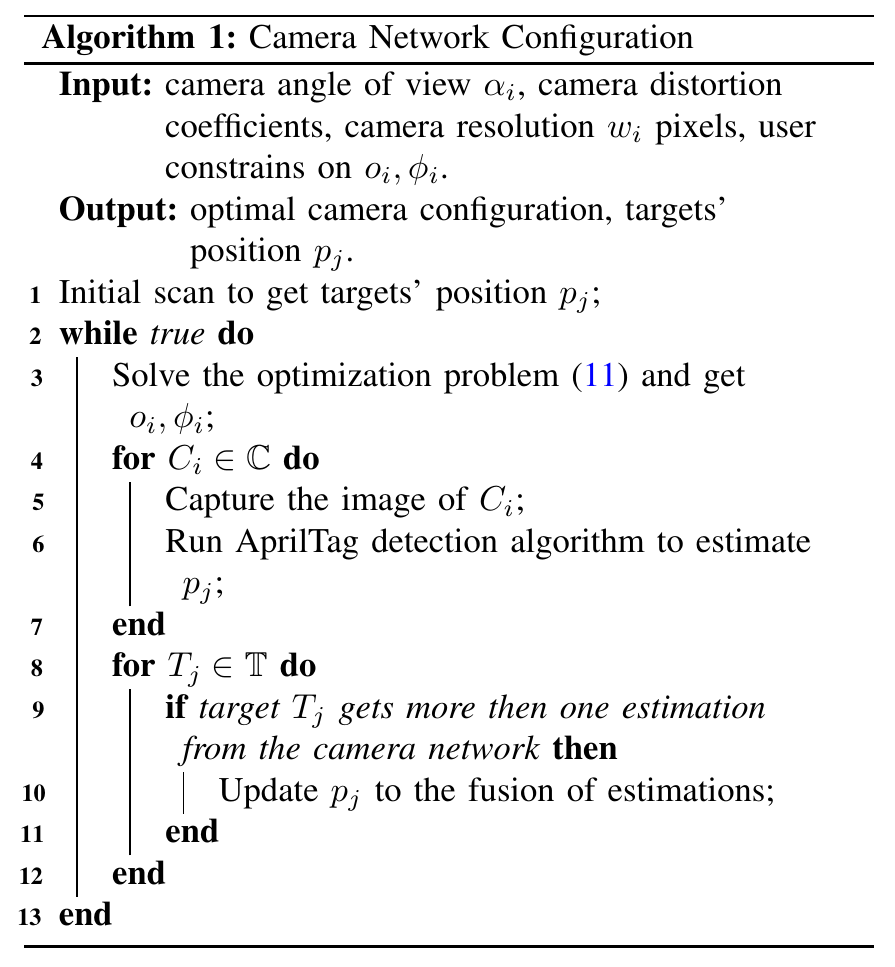}
    % \caption{Caption}
    % \label{fig:my_label}
\end{figure}

\subsection{Optimization}

The tasks of a camera network in the real life vary greatly, resulting in different optimization problems. One of the most common goals is that each target is covered by the field of view of at least one camera and the average sensing quality of all targets is maximized. The objective function based on the proposed sensing quality metric \eqref{eq:fusion result} can be written as
\begin{equation}
\begin{aligned}
    \mbox{Minimize} \quad &\frac{1}{N_t}\sum_{T_j\in \mathbb{T}} \overline{Q}_j\\
    \mbox{subject to} \quad &\sum_{C_i \in \mathbb{C}} V_j^i \geq 1, \forall T_j \in \mathbb{T}.
    % \mbox{subject to} \quad &\prod_{T_j \in \mathbb{T}} \sum_{C_i \in \mathbb{C}} V_j^i > 0.
\end{aligned}
\label{eq:optimization problem}
\end{equation}

Further, with the formulation of sensing quality, different optimization problems can be constructed to deal with other application requirements.
For example, we can optimize the worst-case sensing quality as
\begin{equation}\label{eq:Minmax}
\begin{aligned}
    \mbox{Minimize} \quad &\max_{T_j \in \mathbb{T}} \overline{Q}_j\\
    \mbox{subject to} \quad &\sum_{C_i \in \mathbb{C}} V_j^i \geq 1, \forall T_j \in \mathbb{T}.
    % \mbox{subject to} \quad &\prod_{T_j \in \mathbb{T}} \sum_{C_i \in \mathbb{C}} V_j^i > 0.F
\end{aligned}
\end{equation}

Based on \eqref{eq:optimization problem}, Algorithm 1 is proposed for camera network configuration systems.
Apparently, Algorithm 1 can be used to solve \eqref{eq:Minmax} with the modification of Line 3.

% \begin{algorithm}[h]
%     \LinesNumbered
%     \caption{Camera Network Configuration}
%     \label{alg}
%     \KwIn{camera angle of view $\alpha_i$, camera distortion coefficients, camera resolution $w_i$ pixels, user constrains on $o_i, \phi_i$.
%     }
%     \KwOut{optimal camera configuration, targets' position $p_j$.	
%     }
%     Initial scan to get targets' position $p_j$\;
%     \While{true}
%     {
%         Solve the optimization problem \eqref{eq:optimization problem} and get $o_i, \phi_i$\;
%     	\For{$C_i \in \mathbb{C}$}{    
%         	Capture the image of $C_i$\;
%         	Run AprilTag detection algorithm to estimate $p_j$\;
%             }
%         \For{$T_j \in \mathbb{T}$}{
%             \If{target $T_j$ gets more then one estimation from the camera network}{Update $p_j$ to the fusion of estimations;
%                 }
%             }
%     }
% \end{algorithm}
	
% \begin{figure}[htb]
% 	\centering
% 	\includegraphics[width=0.9\linewidth]{pic/model_flow.png}
% 	\caption{Camera Network Reconfiguration System Overview{\color{red}(useful?)}}
%     \label{fig:System Overview}
% \end{figure}

\section{Performance Evaluation}\label{sec:Experimental Validation}

% 仿真和实验 来证明有效性
\subsection{Simulation}

\begin{figure*}[t]
	\centering
	\begin{subfigure}{0.93\linewidth}
	    
    	\begin{subfigure}{0.24\linewidth}
    		\centering
    		\includegraphics[width=0.9\linewidth]{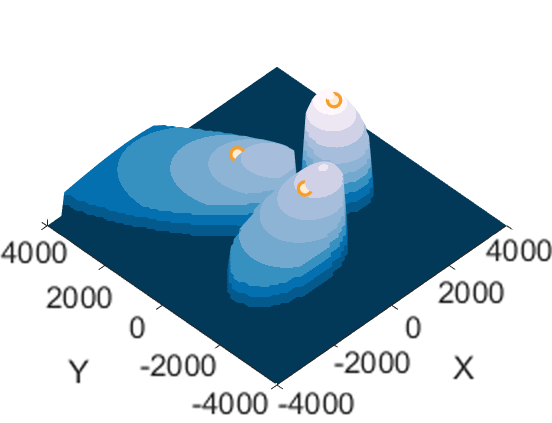}
    		\caption{PTZ, $N_c=3, N_t=3$}
    % 		\label{figure} % 小图标题
    	\end{subfigure}
    	\begin{subfigure}{0.24\linewidth}
    		\centering
    		\includegraphics[width=0.9\linewidth]{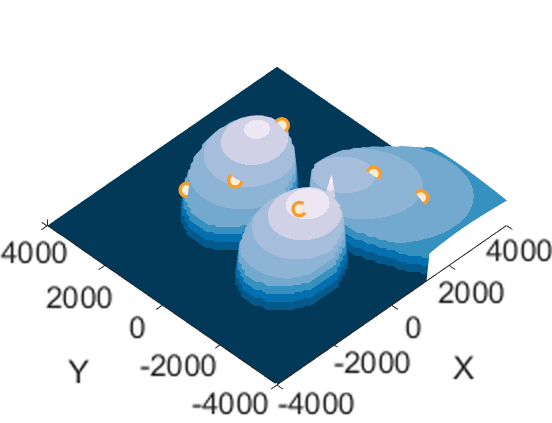}
    		\caption{PTZ, $N_c=3, N_t=9$}
    	\end{subfigure}
    	\begin{subfigure}{0.24\linewidth}
    		\centering
    		\includegraphics[width=0.9\linewidth]{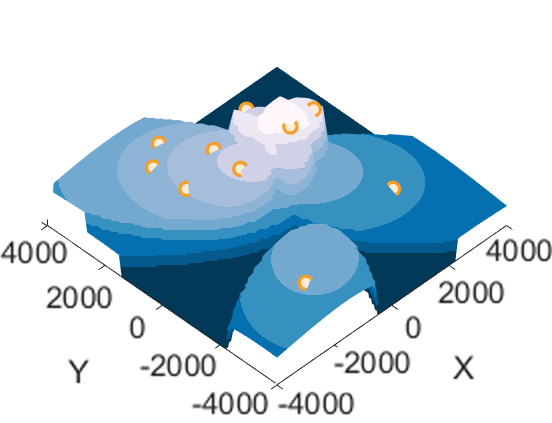}
    		\caption{PTZ, $N_c=5, N_t=10$}
    	\end{subfigure}
    	\begin{subfigure}{0.24\linewidth}
    		\centering
    		\includegraphics[width=0.9\linewidth]{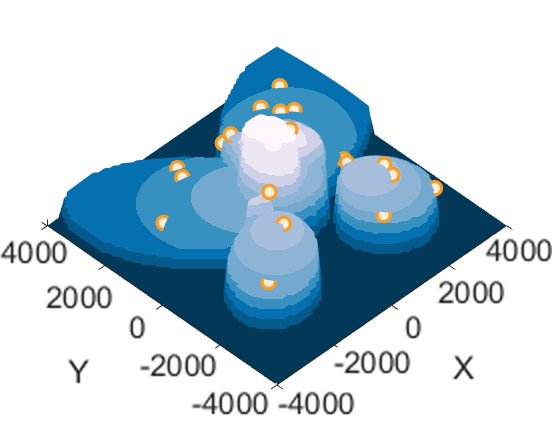}
    		\caption{PTZ, $N_c=5, N_t=20$}
    	\end{subfigure}
    	\quad
    	\begin{subfigure}{0.24\linewidth}
    		\centering
    		\includegraphics[width=0.9\linewidth]{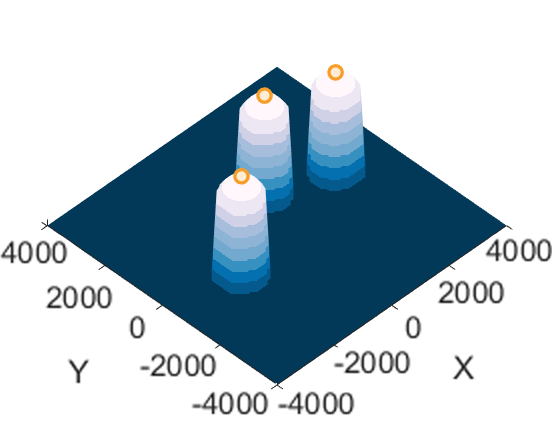}
    		\caption{Drone, $N_c=3, N_t=3$}
    % 		\label{figure} % 小图标题
    	\end{subfigure}
    	\begin{subfigure}{0.24\linewidth}
    		\centering
    		\includegraphics[width=0.9\linewidth]{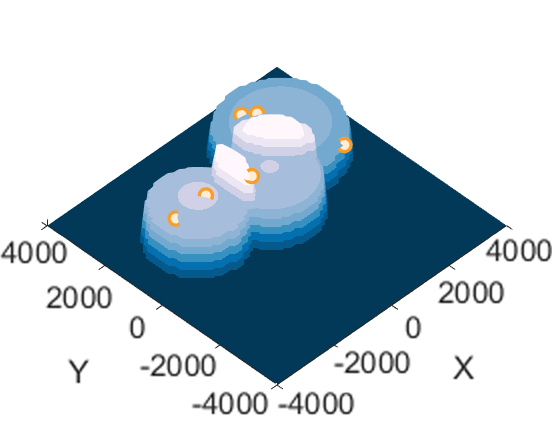}
    		\caption{Drone, $N_c=3, N_t=9$}
    	\end{subfigure}
    	\begin{subfigure}{0.24\linewidth}
    		\centering
    		\includegraphics[width=0.9\linewidth]{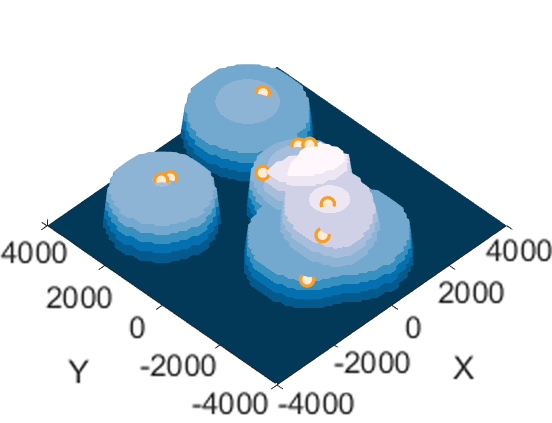}
    		\caption{Drone, $N_c=5, N_t=10$}
    	\end{subfigure}
    	\begin{subfigure}{0.24\linewidth}
    		\centering
    		\includegraphics[width=0.9\linewidth]{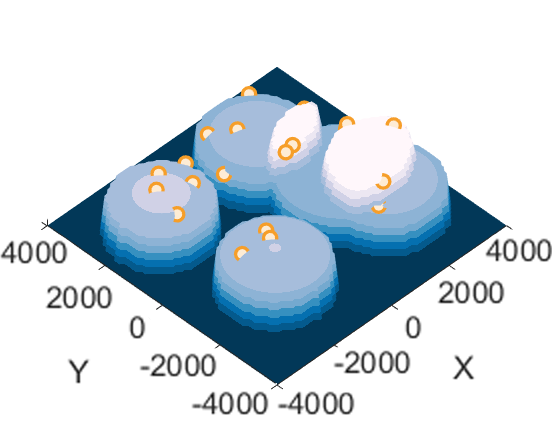}
    		\caption{Drone, $N_c=5, N_t=20$}
    	\end{subfigure}
    	
	\end{subfigure}
	\begin{subfigure}{0.05\linewidth}
		\centering
		\includegraphics[width=0.9\linewidth]{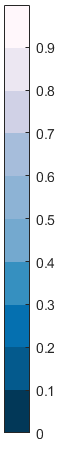}
% 		\caption{drone, 5c20r}
	\end{subfigure}
	
	\caption{Simulations for different network scales. The XY axis represents the experiment plane. The Z axis represents the sensing quality, which has been normalized for better presentation. The targets' positions are marked with circles.
	}
	\label{fig:MATLAB_exp}
	\vspace{-10pt}
\end{figure*}

In this section, we provide simulation experiments demonstrating the claimed strength of proposed sensing quality model for the reconfiguration of heterogeneous camera networks, e.g., PTZ camera networks or drone mode (cameras carried by unmanned aerial vehicles). In order to evaluate the sensing quality, we build a virtual environment to simulate all the camera views. In the designed experiment, several robots move randomly on the 5m $\times$ 3m floor\cite{ding2021robopheus,wang2022safety}. Each robot is attached with an individual visual fiducial marker AprilTag\cite{olson_apriltag_2011}, which facilitates the vision-based six degrees of freedom robot localization. The task of the camera network is to obtain the positions of all the robots as accurately as possible.
\begin{figure}[t]
    \centering
    \includegraphics[width=0.4\textwidth]{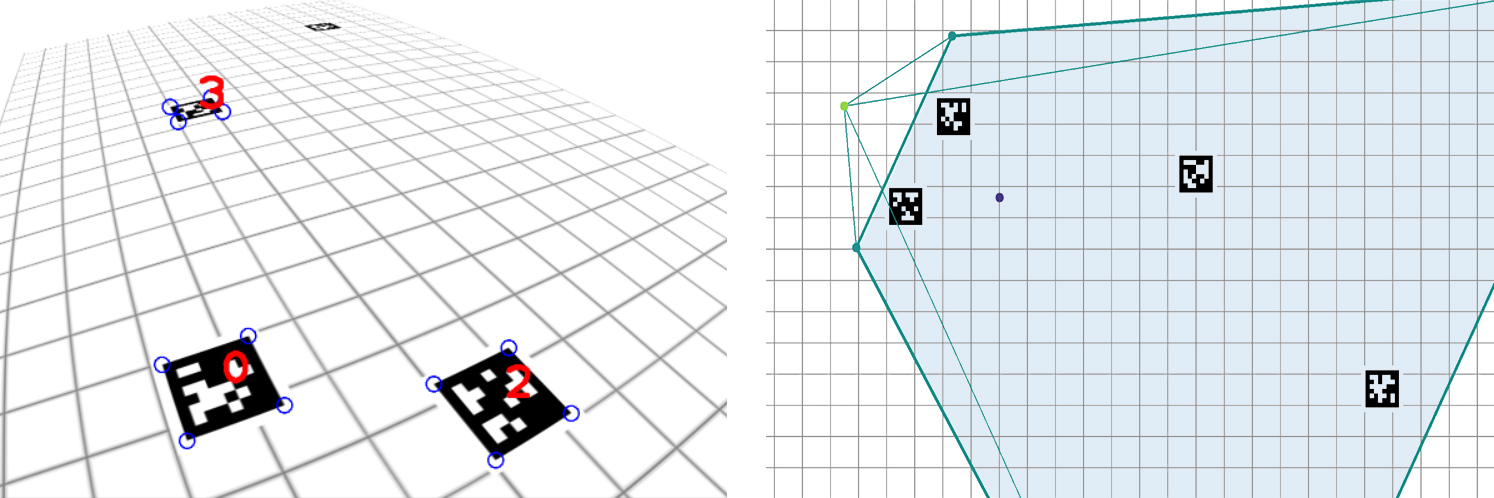}
    \caption{AprilTag detection task in the virtual environment, (left) image space and (right) world space. The farthest tag was not detected successfully because it occupied few pixels on the image.}
    \label{fig:virtual environment}
\end{figure}
Specifically, the robots' position $p_j$ are randomly given and cameras have fixed mounting position $o_i$ and flexible optical-axis direction $\phi_i$ in the PTZ camera mode, while flexible position $o_i$ (minimum height is limited to 1m) but fixed downward optical-axis direction $\phi_i$ in the drone mode. All the cameras' coefficients are calibrated from real cameras. Our virtual environment generates images corresponding to the given camera configuration, shown in Fig. \ref{fig:virtual environment}. AprilTag detection algorithm is used to estimate the position of robots according to the images. The optimization problem \eqref{eq:optimization problem} is solved by \texttt{fmincon} in MATLAB 2021, which runs on an AMD Ryzen 5 3600 CPU, 3.60GHz processor and 32GB RAM. We generate 12 different simulation scenarios and run 10 experiments in each scenario. The results are shown in Fig. \ref{fig:MATLAB_exp} and Table \ref{table: simulation}.

\begin{table*}[t]
\centering
\caption{Simulation Results on Multi-Robot Detection}
\label{table: simulation}
\begin{threeparttable}
\begin{tabular}{ccccccccc}
\toprule[1.5pt]
\multirow{2}{*}{$N_c$} & \multicolumn{1}{l}{\multirow{2}{*}{$N_t$}} & \multicolumn{3}{c}{PTZ}                &  & \multicolumn{3}{c}{Drone}              \\ 
\cline{3-5} \cline{7-9} 
\specialrule{0em}{1pt}{1pt}
                       & \multicolumn{1}{l}{}                       & Method   & Computing Time/s & Error/mm &  & Method   & Computing Time/s & Error/mm \\
\midrule[1pt]
\multirow{3}{*}{3}     & \multirow{3}{*}{3}                         & IPOPT    & 0.684            & 11.97    &  & IPOPT    & 0.534            & 5.13     \\
                       &                                            & SQP      & 0.223            & 11.85    &  & SQP      & 0.120            & 5.67     \\
                       &                                            & Arslan's & -       & 13.56    &  & Arslan's & -                & 9.71     \\ \cline{3-5} \cline{7-9} 
                       \specialrule{0em}{2pt}{2pt}
\multirow{3}{*}{5}     & \multirow{3}{*}{10}                        & IPOPT    & 1.065            & 21.96    &  & IPOPT    & 1.321            & 14.62    \\
                       &                                            & SQP      & 0.490            & 25.69    &  & SQP      & 0.586            & 17.51    \\
                       &                                            & Arslan's & -                & 47.11    &  & Arslan's & -                & 22.18    \\ \cline{3-5} \cline{7-9} 
                       \specialrule{0em}{2pt}{2pt}
\multirow{3}{*}{7}     & \multirow{3}{*}{20}                        & IPOPT    & 1.492            & 23.55    &  & IPOPT    & 1.915            & 17.44    \\
                       &                                            & SQP      & 0.720            & 27.23    &  & SQP      & 0.746            & 21.99    \\
                       &                                            & Arslan's & -                & 51.02    &  & Arslan's & -                & 22.93    \\
\bottomrule[1.5pt]                       
\end{tabular}
\begin{tablenotes}
        \item[*] Arslan's method \cite{arslan_voronoi-based_2018} did not give comparisons on computing time. 
      \end{tablenotes}    
      \end{threeparttable}
\end{table*}

First, the average errors of the PTZ camera systems are larger than that of the drone camera systems. This is because the fixed positions of the PTZ cameras lead to more serious distortion of the sight when monitoring robots at a large relative angle, while drones can fly directly above the robots to get a sight with better sensing quality. 
Besides, we compare the performance of the two algorithms, i.e., Interior Point OPTimizer (IPOPT) and Sequential Quadratic Programming (SQP). The results show that SQP computes faster with larger error.
Moreover, the sensing quality model in \cite{arslan_voronoi-based_2018} is used for comparison. It is obvious that our model achieves more accurate localization since our model incorporates the effects of distortion.
Lastly, as the numbers of cameras and targets increase, the average computing time increases too due to the increasing complexity of the optimization problem, and it becomes harder to find a global optimal solution.

% 初稿版本
% \begin{table*}[tb]
% \centering
% \caption{Simulation Results on Multi-Robot Detection}
% \label{table: simulation}
% \begin{tabular}{ccccccc}
% \hline
% Camera Configuration & $N_c$ & $N_t$ & Average Computing Time/s & Average $\overline{Q}$ & Average Error by Ours/mm & Average Error by Arslan's\cite{arslan_voronoi-based_2018}/mm \\ \hline
% PTZ camera & 3 & 3 & 0.615 & 0.664 & 13.76 & 13.56 \\
% PTZ camera & 3 & 6 & 1.281 & 1.207 & 23.84 & 52.84 \\
% PTZ camera & 5 & 10 & 1.417 & 1.258 & 24.71 & 56.03 \\
% PTZ camera & 5 & 20 & 1.965 & 1.972 & 40.87 & 75.62 \\
% drone camera & 3 & 3 & 1.214 & 0.147 & 3.94 & 8.51 \\
% drone camera & 3 & 6 & 1.393 & 0.991 & 18.92 & 23.51 \\
% drone camera & 5 & 10 & 1.719 & 0.926 & 17.68 & 24.84 \\
% drone camera & 5 & 20 & 1.975 & 1.372 & 23.98 & 39.72 \\ \hline
% \end{tabular}
% \end{table*}

\subsection{Experiment}

\begin{figure*}[t]
    \centering
	\begin{subfigure}{0.51\linewidth}
		\centering
		\includegraphics[width=0.9\linewidth]{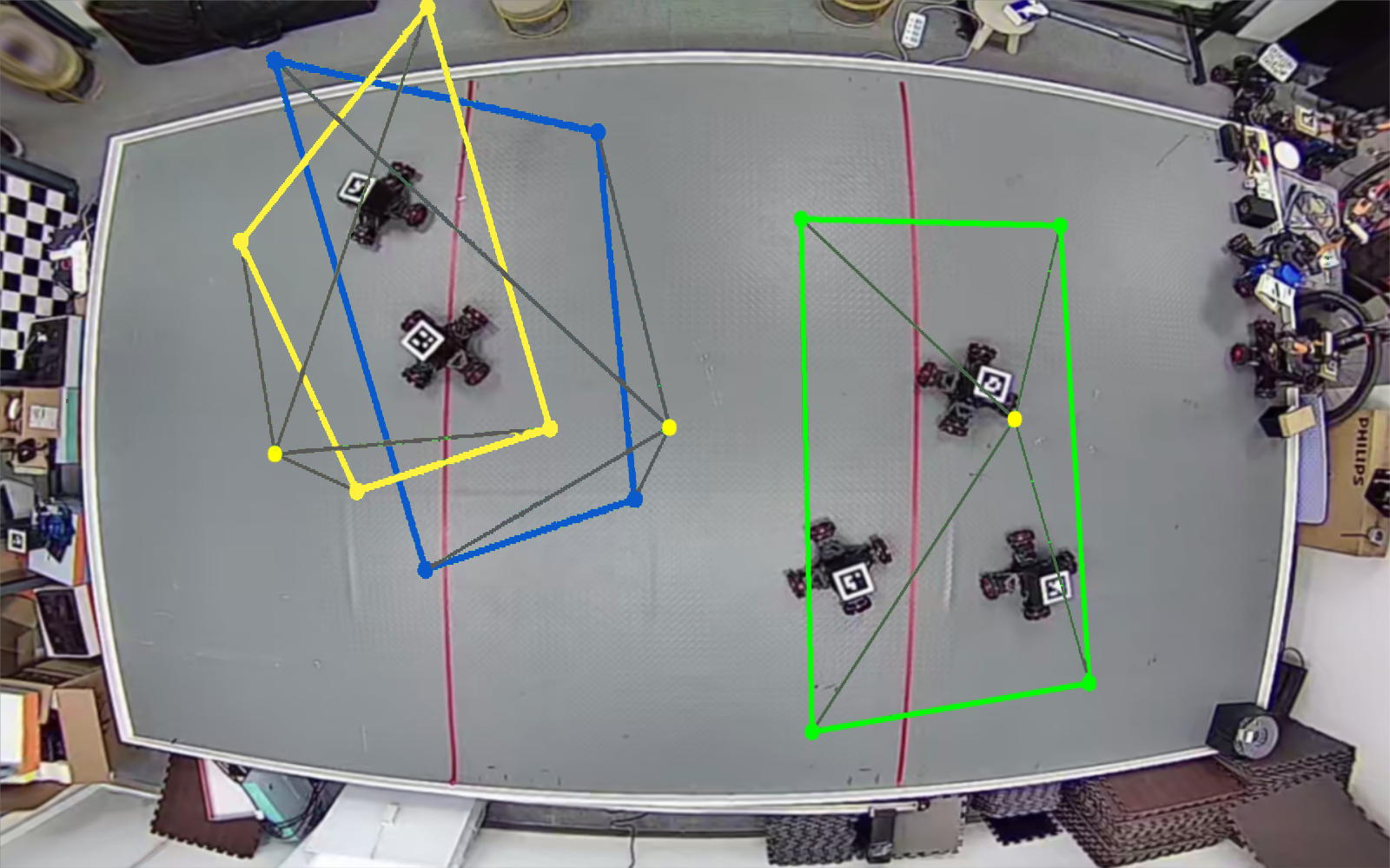}
	\end{subfigure}
	\begin{subfigure}{0.47\linewidth}
		\centering
		\begin{subfigure}{0.47\linewidth}
    		\centering
    		\includegraphics[width=0.9\linewidth]{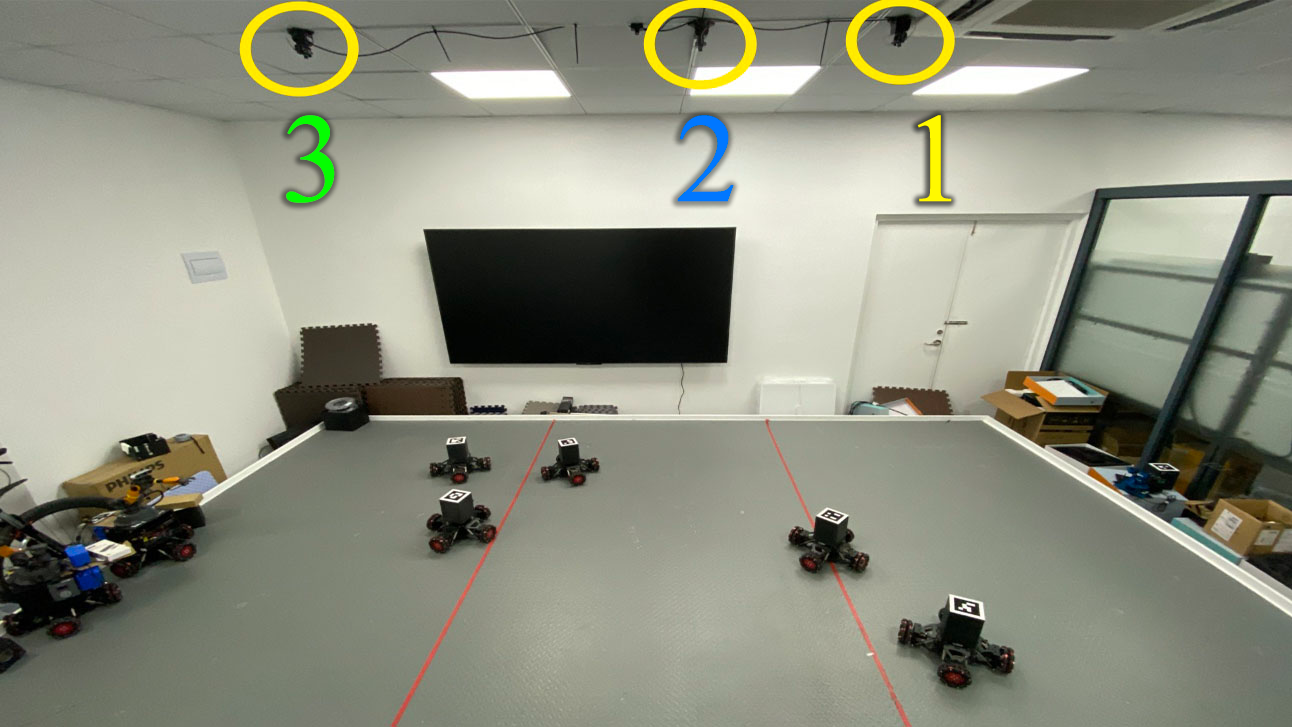}
    		\caption{Experiment Overview}
    	\end{subfigure}
		\begin{subfigure}{0.48\linewidth}
    		\centering
    		\includegraphics[width=0.9\linewidth]{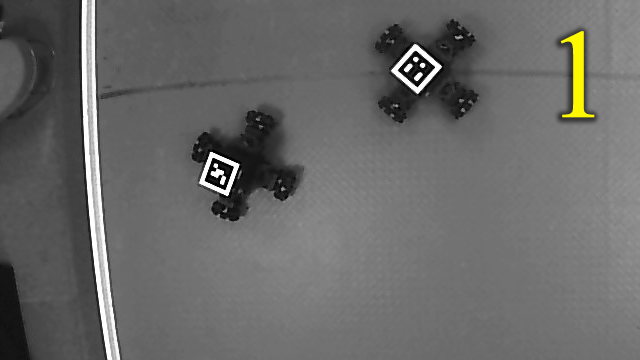}
    		\caption{Camera 1 View}
    	\end{subfigure}
    	\quad
		\begin{subfigure}{0.48\linewidth}
    		\centering    
    		\includegraphics[width=0.9\linewidth]{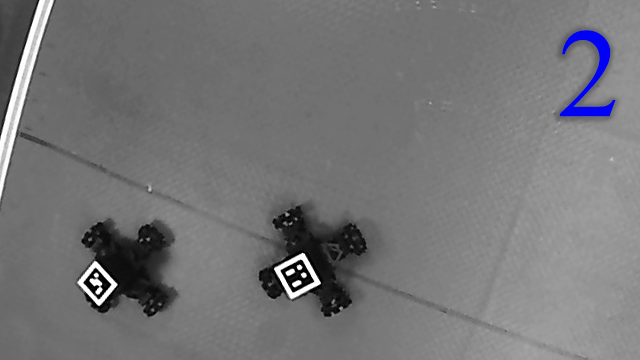}
    		\caption{Camera 2 View }
    	\end{subfigure}
		\begin{subfigure}{0.48\linewidth}
    		\centering
    		\includegraphics[width=0.9\linewidth]{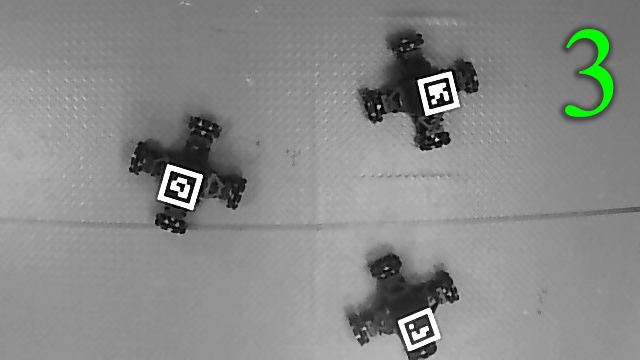}
    		\caption{Camera 3 View}
    	\end{subfigure}
	\end{subfigure}
    \caption{Experimental demonstration of camera network configuration for optimal sensing quality. (left) Top view of the environment and visualization of a optimal configuration for certain situation. (a) Three PTZ cameras mounted on the ceiling. (b)(c)(d) The camera view with the optimized configuration. }
    \label{fig:exp_result}
    \vspace{-10pt}
\end{figure*}

To experimentally validate the proposed model, we set up a 3m (width) $\times$ 5m (length) $\times$ 2.5m (height) rectangular box-shaped environment with three self-build interchangeable-lens PTZ cameras mounted on the ceiling, the intrinsic and extrinsic coefficients of which are pre-calibrated. We set several Omni-directional robots attached with AprilTags moving on the ground as sensing targets. The cameras' positions are fixed, and the optical-axis directions perform as optimization variables, limited by PTZ cameras' physical properties. Since the estimation of the depth depends on the size of the tags in the image, this experiment can examine both the positioning and measurement ability simultaneously. 
% At the beginning of the experiment, a fisheye camera installed above the center of the platform performs a global scan to send out the initial position of the robot.
The ground truth of the robots' positions is given by manual measurement with laser rangefinder. Our algorithm will give an optimal camera configuration for sensing quality, as shown in Fig. \ref{fig:exp_result} and \href{https://youtu.be/5Uies1RrC-0}{video}. 

We first use manual configuration of the cameras to cover all the robots and get the average error. Then, the configuration are set according to Algorithm 1. The average $\overline{Q}$ is 1.165 and average error decreased by 37.0\%, down to 35.3.
Finally, we use sensing model proposed by \cite{arslan_voronoi-based_2018} as another comparison. The results are shown in Table \ref{table: experiment}.

\begin{table}[b]
\centering
\caption{Experimental Results on Multi-Robot Detection}
\label{table: experiment}
\begin{tabular}{ccc}
\hline
Method & Average $\overline{Q}$ & Average Error/mm\\ \hline
Manual Setting & 1.610 & 56.0\\
Our Method & \textbf{1.165} & \textbf{35.3}\\
Arslan's\cite{arslan_voronoi-based_2018} & 1.482 & 37.2\\ \hline
\end{tabular}
\end{table}

\section{Conclusion}\label{sec:Conclusion}
% 贡献：新问题、新模型、实验
% 迁移能力 + Future: 合适的求解方法

In this paper, we considered the balance between the coverage for all targets and sensing quality of each single one for a camera network. The problem has been formulated as an optimization problem which takes into account the perspective, distortion factor and multi-camera fusion results. Optimal configuration for camera network can be achieved without modeling the environment, suitable for any pre-calibrated cameras and targets anywhere in 3D space. The proposed method demonstrates superior performance as compared to previous approaches, increasing the average accuracy in AprilTag detection task by $38.9\%$.

We would like to introduce some potential research directions: i) developing optimization approach for online tracking and large-scale problems, ii) optimizing sensing quality while taking the surveillance of unknown areas into account, and iii) developing a refined model for objects which need to be tracked from a certain direction.
% maybe it's called anisotropic

\balance
\bibliographystyle{ieeetr}
\bibliography{main}

\begin{thebibliography}{10}

\bibitem{chen2021automated}
C.~Chen, R.~Surette, and M.~Shah, ``Automated monitoring for security camera
  networks: promise from computer vision labs,'' {\em Security Journal},
  vol.~34, no.~3, pp.~389--409, 2021.

\bibitem{bernhard2021case}
C.~Bernhard, R.~Reinhard, M.~Kleer, and H.~Hecht, ``A case for raising the
  camera: a driving simulator test of camera-monitor systems,'' {\em Human
  Factors}, p.~00187208211010941, 2021.

\bibitem{wu2018image}
Y.~Wu, F.~Tang, and H.~Li, ``Image-based camera localization: an overview,''
  {\em Visual Computing for Industry, Biomedicine, and Art}, vol.~1, no.~1,
  pp.~1--13, 2018.

\bibitem{yang2020multi}
Y.~Yang, D.~Tang, D.~Wang, W.~Song, J.~Wang, and M.~Fu, ``Multi-camera visual
  slam for off-road navigation,'' {\em Robotics and Autonomous Systems},
  vol.~128, p.~103505, 2020.

\bibitem{yang2017multi}
S.~Yang, S.~A. Scherer, X.~Yi, and A.~Zell, ``Multi-camera visual slam for
  autonomous navigation of micro aerial vehicles,'' {\em Robotics and
  Autonomous Systems}, vol.~93, pp.~116--134, 2017.

\bibitem{zhang2018visual}
X.~Zhang, X.~Chen, F.~Farzadpour, and Y.~Fang, ``A visual distance approach for
  multicamera deployment with coverage optimization,'' {\em IEEE/ASME
  Transactions on Mechatronics}, vol.~23, no.~3, pp.~1007--1018, 2018.

\bibitem{suresh2020maximizing}
M.~S. Suresh, A.~Narayanan, and V.~Menon, ``Maximizing camera coverage in
  multicamera surveillance networks,'' {\em IEEE Sensors Journal}, vol.~20,
  no.~17, pp.~10170--10178, 2020.

\bibitem{horster2006optimal}
E.~H{\"o}rster and R.~Lienhart, ``On the optimal placement of multiple visual
  sensors,'' in {\em Proceedings of the 4th ACM International Workshop on Video
  Surveillance and Sensor Networks}, pp.~111--120, 2006.

\bibitem{mavrinac_modeling_2013}
A.~Mavrinac and X.~Chen, ``Modeling {Coverage} in {Camera} {Networks}: {A}
  {Survey},'' {\em International Journal of Computer Vision}, vol.~101,
  pp.~205--226, Jan. 2013.

\bibitem{7095560}
C.~Piciarelli, L.~Esterle, A.~Khan, B.~Rinner, and G.~L. Foresti, ``Dynamic
  reconfiguration in camera networks: A short survey,'' {\em IEEE Transactions
  on Circuits and Systems for Video Technology}, vol.~26, no.~5, pp.~965--977,
  2016.

\bibitem{o1987art}
J.~O'rourke, {\em Art gallery theorems and algorithms}, vol.~57.
\newblock Oxford New York, NY, USA, 1987.

\bibitem{efrat2000sweeping}
A.~Efrat, L.~J. Guibas, S.~Har-Peled, D.~C. Lin, J.~S. Mitchell, and T.~Murali,
  ``Sweeping simple polygons with a chain of guards,'' in {\em Proceedings of
  the Eleventh Annual ACM-SIAM Symposium on Discrete Algorithms}, pp.~927--936,
  2000.

\bibitem{konda_optimal_2013}
K.~R. Konda and N.~Conci, ``Optimal configuration of {PTZ} camera networks
  based on visual quality assessment and coverage maximization,'' in {\em 2013
  {Seventh} {International} {Conference} on {Distributed} {Smart} {Cameras}
  ({ICDSC})}, pp.~1--8, IEEE, Oct. 2013.

\bibitem{konda_global_2016}
K.~R. Konda, N.~Conci, and F.~De~Natale, ``Global {Coverage} {Maximization} in
  {PTZ}-{Camera} {Networks} {Based} on {Visual} {Quality} {Assessment},'' {\em
  IEEE Sensors Journal}, vol.~16, pp.~6317--6332, Aug. 2016.

\bibitem{aghajanzadeh_camera_2020}
S.~Aghajanzadeh, R.~Naidu, S.-H. Chen, C.~Tung, A.~Goel, Y.-H. Lu, and G.~K.
  Thiruvathukal, ``Camera {Placement} {Meeting} {Restrictions} of {Computer}
  {Vision},'' in {\em 2020 {IEEE} {International} {Conference} on {Image}
  {Processing} ({ICIP})}, pp.~3254--3258, IEEE, Oct. 2020.

\bibitem{arslan_voronoi-based_2018}
O.~Arslan, H.~Min, and D.~E. Koditschek, ``Voronoi-based coverage control of
  pan/tilt/zoom camera networks,'' in {\em 2018 {IEEE} International Conference
  on Robotics and Automation ({ICRA})}, pp.~5062--5069, {IEEE}.

\bibitem{kuroda2017essential}
T.~Kuroda, {\em Essential principles of image sensors}.
\newblock CRC press, 2017.

\bibitem{robinson1988optical}
I.~Robinson and A.~Trautman, ``Optical geometry,'' in {\em New Theories in
  Physics}, 1988.

\bibitem{bisagno_virtual_2018}
N.~Bisagno and N.~Conci, ``Virtual camera modeling for multi-view simulation of
  surveillance scenes,'' in {\em 2018 26th {European} {Signal} {Processing}
  {Conference} ({EUSIPCO})}, pp.~2170--2174, IEEE, Sept. 2018.

\bibitem{fryer1986lens}
J.~G. Fryer and D.~C. Brown, ``Lens distortion for close-range
  photogrammetry,'' {\em Photogrammetric Engineering and Remote Sensing},
  vol.~52, pp.~51--58, 1986.

\bibitem{zhang1999flexible}
Z.~Zhang, ``Flexible camera calibration by viewing a plane from unknown
  orientations,'' in {\em Proceedings of the Seventh IEEE International
  Conference on Computer Vision}, vol.~1, pp.~666--673, IEEE, 1999.

\bibitem{ding2021robopheus}
X.~Ding, H.~Wang, H.~Li, H.~Jiang, and J.~He, ``Robopheus: A virtual-physical
  interactive mobile robotic testbed,'' {\em arXiv preprint arXiv:2103.04391},
  2021.

\bibitem{wang2022safety}
H.~Wang, X.~Ding, J.~He, K.~Margellos, and A.~Papachristodoulou, ``Safety-aware
  optimal control in motion planning,'' {\em arXiv preprint arXiv:2204.13380},
  2022.

\bibitem{olson_apriltag_2011}
E.~Olson, ``{AprilTag}: {A} robust and flexible visual fiducial system,'' in
  {\em 2011 {IEEE} {International} {Conference} on {Robotics} and {Automation}
  ({ICRA})}, pp.~3400--3407, IEEE, May 2011.

\end{thebibliography}
\end{document}